\documentclass[twoside,11pt]{article}

\usepackage{jmlr2e}
\usepackage{booktabs}

\ShortHeadings{An autoencoder for causal discovery}{Feiler}
\firstpageno{1}

\begin{document}

\title{A probabilistic autoencoder for causal discovery}

\author{\name Matthias J. Feiler \email matthias.feiler@uzh.ch \\
	\addr Department of Banking and Finance\\
	University of Zurich\\
	CH-8032 Zürich, Switzerland
}

\editor{}

\maketitle

\begin{abstract}
The paper addresses the problem of finding the causal direction between two associated variables. The proposed solution is to build an autoencoder of their joint distribution and to maximize its estimation capacity relative to both the marginal distributions. It is shown that the resulting two capacities cannot, in general, be equal. This leads to a new criterion for causal discovery: the higher capacity is consistent with the unconstrained choice of a distribution representing the cause while the lower capacity reflects the constraints imposed by the mechanism on the distribution of the effect. Estimation capacity is defined as the ability of the auto-encoder to represent arbitrary datasets. A regularization term forces it to decide which one of the variables to model in a more generic way i.e., while maintaining higher model capacity. The causal direction is revealed by the constraints encountered while encoding the data instead of being measured as a  property of the data itself. The idea is implemented and tested using a restricted Boltzmann machine.
\end{abstract}


\section{Introduction}

Causal discovery is the problem of identifying the causal structure (in the form of a directed acyclic graph) from observational data. Established techniques search for conditional independence relations in the data \citep{spirtes2000causation}, \citep{pearl2009causality}, \citep{glymour2019review} and reside on two assumptions: first, the joint distribution is \textit{Markov} with respect to the causal graph i.e., each variable is independent of all its non-descendants given its parents. Second, it is \textit{faithful} i.e., all conditional independences are due to the Markov condition. Constraint based methods are able to identify the causal graph up to Markov equivalence. Such methods fail in the case of two variables $X$ and $Y$ since $X \rightarrow Y$ and $X \leftarrow Y$ are Markov equivalent. In this paper, a new method for causal discovery of bivariate data is proposed.

Following Reichenbach's principle \citep{reichenbach1991direction}, all statistical dependency is due to some physical mechanism that connects the variables in question. This can be modeled by postulating that nature first selects a cause $X$ and then independently a mechanism that transforms the cause into effect $Y$. The resulting structural causal model (SCM) entails a joint distribution $P_{X, Y}$. $P_{X, Y}$, in turn, does \textit{not} define the SCM as every joint distribution admits an SCM in both directions. Additional assumptions are required to make the structure identifiable, see \citep{peters2017elements} for an introduction to this fundamental fact. One type of assumption is to restrict the model classes describing cause and mechanism. A number of restrictions have been proposed including linear models with non-Gaussian additive noise \citep{kano2003causal}, the nonlinear additive noise model \citep{hoyer2008nonlinear} as well as the post-nonlinear causal model \citep{zhang2012identifiability}. Another type of assumption formalizes the idea of independence of cause and mechanism. Here, independence is understood in terms of algorithmic information. The shortest program that prints a sequence and then halts is the \textit{Kolmogorov complexity} of the sequence. \citep{lemeire2006causal}, \citep{janzing2010causal} develop the viewpoint that if $X \rightarrow Y$ in a causal graph then $P_X$ and $P_{Y|X}$ are algorithmically independent in the sense that knowledge of $P_X$ does not permit a shorter description of $P_{Y|X}$ and vice versa. \citep{janzing2010justifying} use this definition to argue that if $P_Y$ and $P_{X|Y}$ share enough algorithmic information (on which they derive a lower bound) they are tuned to each other in a way that $Y\rightarrow X$ can be rejected.

As the description length in the ideal (Kolmogorov) sense is uncomputable, testable criteria need to be found. The objective of this paper is to introduce a new criterion for the algorithmic independence of $P_X$ and $P_{Y|X}$. Our starting point and benchmark is the information-geometric criterion (IGCI) which formalizes independence via orthogonality in information space \citep{janzing2012information}. If $X$ and $Y$ are related by a deterministic invertible function $f$ then independence of $P_X$ and $f$ implies dependence of $P_Y$ and $f^{-1}$. Here, $P_X$ and $f$ are independent if the covariance of the density $p_X$ and log-derivative $\log f'$, both seen as random variables defined on $[0, 1]$, vanishes. An important insight is that if $P_X$ is uniform, then $P_Y$ reveals the constraints imposed by $f$ on the pair $(X, Y)$. 

The viewpoint developed in this paper is as follows: If $X$ is the true cause then it will be easier to remove structural constraints on $X$ than on $Y$ in a model of their joint distribution $P_{X,Y}$.

\section{Outline}

The idea is to represent the joint distribution of cause and effect $P_{X,Y}$ by an auto-encoder (AE) i.e., a neural network that is trained to reproduce observations $(x,y)$ sampled from $P_{X,Y}$. I claim that the causal direction can be discovered by regularizing the model-implied marginals. The AE consists of a finite number of basis functions distributed over the data e.g., Gaussian decoding functions whose modes are determined by the encoder. Regularization constraints on $X$ and $Y$ complement each other in the following sense: The mode locations can be regular in one direction as long as they are adapted in the other such that all data points are covered as illustrated in figure 1. Any constraint on $X$ needs to be complemented by a constraint on $Y$ (and vice versa) to obtain a high data likelihood.   

The location of the modes reveals structural constraints encountered by the AE when assigning likelihoods to data sets. These constraints can be expressed as the narrowness of the model class represented by the AE. If $X$ causes $Y$ then $P_X$ is determined independently of the effect $P_Y$. The distribution of the effect $P_Y$, in turn, is constrained by the cause through a (causal) mechanism. This means that the marginal $P_X$ belongs to a larger model class than $P_Y$. Assume indeed that $P_X$ belongs to some special model class e.g., the set of bi-modal distributions. Then $P_Y$ cannot belong to a larger model class e.g., all multi-modal distributions, unless the mechanism that transforms $P_X$ into $P_Y$ is adapted to $P_X$ in a way that removes its characteristic bi-modality. However, by the independence postulate, the mechanism does not interact with the cause which means that $Y$ is more constrained than $X$. 

The paper proposes a way of comparing the structure constraints on $X$ and $Y$. This is related to the IGCI where irregularities in $X$ and $Y$ are compared by estimating the entropy of their marginal distributions(\cite{janzing2010justifying}). Our approach is different in that it does not \textit{measure} structure but attempts to remove structure in a joint model of both variables and concludes that this attempt will almost always fail for one of them. The joint model is obtained in an unsupervised machine learning framework which provides a an efficient representation of the observed data. It will be seen that a uniform distribution of the decoding functions in one direction is equivalent to maximizing the representation capacity of the AE for the variable corresponding to that direction. It is therefore useful to define: 

\textbf{Definition 1:} The AE model of $P_{X, Y}$ is ``uniform in $X$ (or $Y$)'' if the centers are uniformly distributed in the $X$ (or $Y$) dimension.

In summary, we build a model for the joint distribution and study the constraints this imposes on capacity of the AE to represent classes of \textit{marginal} distributions. If $X$ is the cause, it is less constrained than $Y$ and it will therefore be less costly to achieve uniformity in $X$ than in $Y$. 

Regularization is typically introduced to prevent overfitting and improve the signal-noise separation capability of the AE. Without regularization, the modes of the Gaussians would be placed to optimally cover the observed data. If $X$ is the true cause (i.e., there is no common cause of $X$ and $Y$) its distribution can be chosen arbitrarily and should not be part of a causal model between $X$ and $Y$. The unregularized AE leads to an overfit since the $X$ coordinates of the modes are part of the overall fitting procedure of the \textit{observed} data. Regularization towards uniformity has the effect of keeping the model general in one of the variables. Notice that this does not mean that the marginal of $X$ is itself uniform but merely that a larger number of distributions can be accommodated by the model. Whatever the observed distribution of $X$, the model class associated with $Y$ will be smaller as it is further constrained by the presence of the causal mechanism.

\begin{figure*}[!htb]
	\vskip 0.2in
	\begin{center}
		\centerline{\includegraphics[width=0.85\columnwidth]{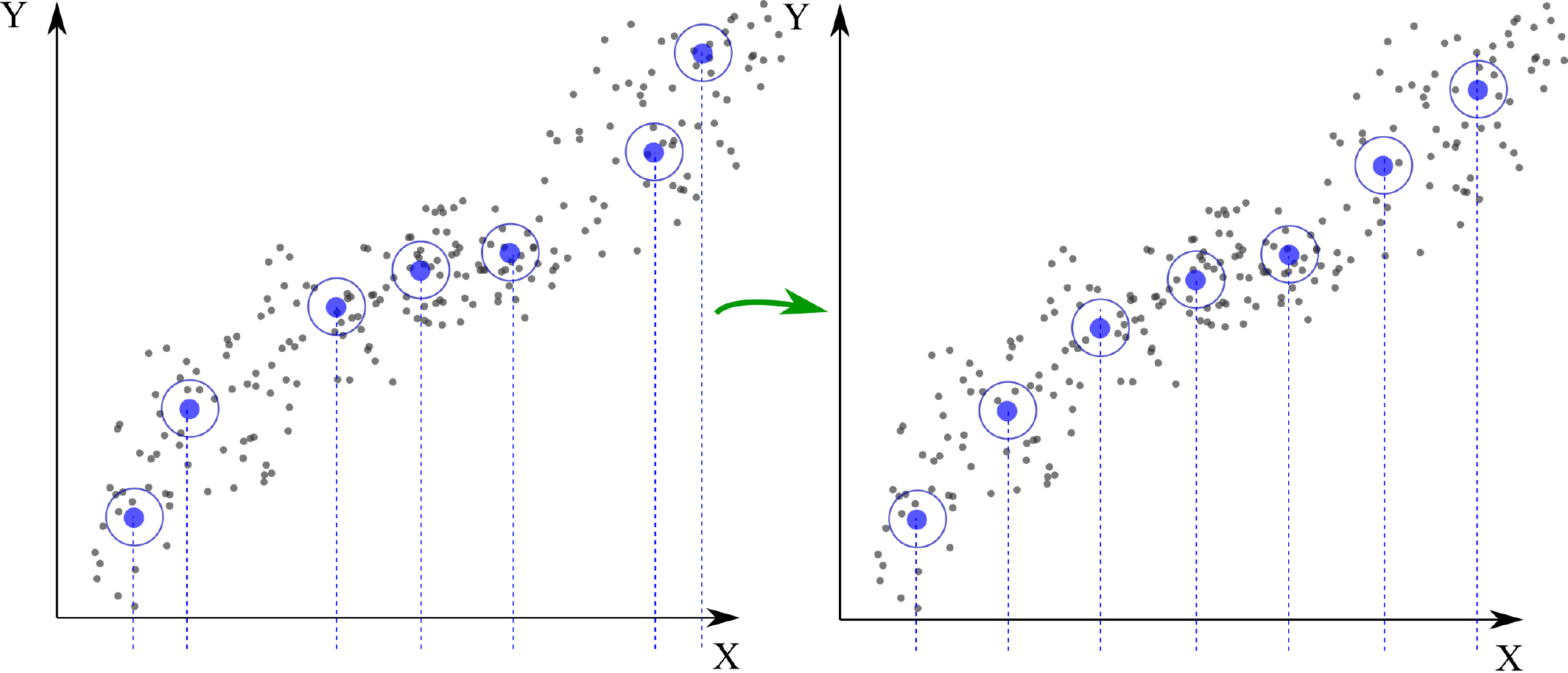}}
		\caption{Regularization yields uniformity of the AE in $X$ (rhs): if $X$ is the true cause, a better signal/noise separation is obtained if all structure constraints on the density $p_X$ are removed.}
		\label{fig:roc}
	\end{center}
	\vskip -0.2in
\end{figure*}

\section{A new route to causal discovery}

This section introduces the building blocks of the proposed method and contains the main result. 

\subsection{Restricted Boltzmann machines}

AEs consist of an encoding and decoding function. We let both functions be stochastic mappings ``$H$ given $V$'' $P_{H|V}$ and ``$V$ given $H$'' $P_{V|H}$, where $V$ refers to the (visible) inputs and $H$ to hidden activations that represent the input. Here, $V=[X, Y]^T$ is a two--dimensional vector combining the cause and effect variables and $H \in \{0,1\}^m$ is an $m$-dimensional vector of binary hidden variables. Both mappings depend on parameters $\theta$ that need to be learned from observations.  

A well-known issue in latent variable models is the intractability of the posterior $P_{H|V}$. A common solution is to minimize a variational lower bound on the reconstruction loss of the AE, see \citep{kingma2019introduction}. Here we prefer to use the restricted Boltzmann machine where $P_{H|V}$ can be evaluated explicitly. 
RBMs are well documented \citep{smolensky1986information}, \citep{hinton2007boltzmann}, \citep{salakhutdinov2007restricted}, so we merely highlight a few important aspects. The training of the RBMs can be motivated by the minimum description length principle. In essence, the aim is to minimize the code-length needed to describe both the hidden and observed variables. The solution is to exploit the regularities in the data such that it can be described with fewer symbols than the number of symbols that would be needed for a literal description. All non-essential data components, including the noise, are discarded as they do not contribute to the performance\footnote{This refers to the out-of-sample reconstruction loss obtained by cross validation.} of the model. It can be shown that the shortest code length is achieved by choosing hidden activations stochastically according to a Boltzmann distribution \citep{hinton1993autoencoders}. Quite often, the hidden layer is also structurally constrained e.g., by letting $H$ be a (low-dimensional) vector of binary variables. 

All functions of the RBM are defined in terms of realizations of $H$ and $V$ so we use the convenient notation for example, $p(h|v)$ instead of $p_{H|V=v}(h)$ when referring to the encoding function. The RBM does not contain any connections among the hidden neurons so we can write  
\begin{equation}
	\label{eq:condi}
	p(h|v) = \textstyle \prod_i p(h_i|v)
\end{equation}
where
\begin{equation}
	\label{eq:fwd}
	p(h_i|v) = h_i \phi[c_i + W_i v] + (1-h_i) (1-\phi[c_i + W_i v]).  
\end{equation}
In (\ref{eq:fwd}), a linear combination of inputs is mapped to the unit interval by means of the sigmoid squashing function $\phi[\cdot]: \mathbb{R} \rightarrow [0,\,1]$. $W_i$ denotes the $i$-th row in the connection matrix $W$ and $c_i$ is the $i$th component of the hidden bias vector, $i=1,\dots,m$. $h_i \in \{0,\,1\}$ is the binary activation of the $i$th hidden unit. In the decoding direction, we have 
\begin{equation}
	\label{eq:vcondi}
	p(v|h) = \textstyle \prod_j p(v_j|h).
\end{equation}
where $v_j$ is the $j$th component of $v$ and 
\begin{equation}
	\label{eq:back}
	p(v_j|h) = \mathcal{N} (v_j |  b_j + W_j^Th, \,\sigma^2). 
\end{equation}
as also the visible nodes are not connected to each other. $\mathcal{N}(v_j|\cdot, \cdot)$ is a Gaussian distribution with mean $b_j + W_j^Th$ and variance $\sigma^2$. In this paper, the input is two-dimensional, so $j=1,2$. The Gaussian-Bernoulli RBM only models the conditional mean of the input given $h$. More sophisticated models exist for capturing also the conditional covariance, see e.g. \citep{courville2011spike}.

The RBM is trained in an unsupervised fashion such as to maximize the (log) likelihood of the data $p(v)$ using gradient descent. A breakthrough in the practical use of RBMs was made by introducing the so-called contrastive divergence (CD) loss which can be computed efficiently. It should be noted that contrastive divergence is by itself not an approximation of (minus) the log likelihood but merely a computational tool for calculating an update direction that approximates the gradient of $\log p(v)$. The CD loss function is defined as the difference between the so-called free energies of the true and reconstructed data points \citep{hinton2002training}.

\subsection{Regularization of the RBM}

The RBM is constrained in two ways: first, the number of activations that can be generated in the hidden layer is bounded above by $2^m$ where $m$ is the number of hidden units. Second, the function class available to the decoder is limited to the family of normal density functions. Both limitations are intended as discussed above. However, we have not yet incorporated our prior knowledge that one of the marginal distributions is subject to less structure constraints than the other. 

Let $x^0$ denote the set of (given) observations and let $x^i$, $i > 0$ be alternative observation sets. $\mathcal{X} = \cup_{i\geq 0} x^i$ contains all conceivable sets of realizations of $X$ (including the given data). Similarly, $\mathcal{Y}$ is the set of all possible realizations of $Y$. Furthermore, let $p(\mathcal{X})$ or $p(\mathcal{Y})$ be the likelihoods assigned by the AE model to the observation sets.F

\textbf{Definition 2:} The AE model of $P_{X, Y}$ is said to have less structure in $X$ than in $Y$ if $p(\mathcal{X}) > p(\mathcal{Y})$.

The AE is trained to assign high data likelihood to \textit{pairs} of observations $\{(x,y)\}$. Without regularization, the inequality can hold in either direction since the model does not include any objectives regarding the marginal distributions. The regularization idea is to maximize both $p(X)$ and $p(Y)$ and find that the achieved maxima are unequal. The criterion in definition 2 cannot be evaluated directly since the alternative observation sets constituting $\mathcal{X}$ or $\mathcal{Y}$ are not given. We use Rissanen's notion of model complexity (sometimes also referred to as its estimation capacity \citep{rissanen2012optimal}) to develop a data-independent alternative criterion.

Let $p(u|\theta)$ be a distribution over some sample space $\mathcal{U}$ and $\theta$ a parameter vector contained in some bounded set $\Omega$ then a ``model class'' refers to the parametric family of distributions 
\begin{equation}
\label{eq:mk}
\mathcal{M} = \{p(u|\theta) \,|\, \theta \subset \Omega\}.
\end{equation}
Furthermore, let $\hat \theta(\cdot): u \rightarrow \hat \theta(u)$ be a function that assigns to every data point $u \in \mathcal{U}$ the parameter that maximizes its likelihood i.e., 
\begin{equation}
\label{eq:ek}
\hat \theta(u) = \mbox{argmax}_\theta p(u| \theta)
\end{equation}
The estimation capacity for a model class $\mathcal{M}$ is given by $\log C$ where
\begin{equation}
\label{eq:ek2}
C = \int_{\mathcal{U}} p(u| \hat \theta(u)) du
\end{equation}
i.e., it is the logarithm of the area below all \textit{best fits} the model class $\mathcal{M}$ can provide. Interpreting $h$ as a parameter we can define 
\begin{equation}
\label{eq:cT}
C_{RBM} = \int_{\mathcal{U}} p(u|\hat h(u)) du.	
\end{equation}
where $p(u|h)$ is the decoder density (\ref{eq:vcondi}) and (\ref{eq:back}) evaluated at data points $u\in \mathcal{U}$ and $\mathcal{H}$ is a (finite) set of activations. We will be interested in the estimation capacity $C_{RBM,j}$ relative to the $j$th input coordinate $j=1,2$: 
\begin{equation}
\label{eq:cTj}
C_{RBM,j} = \int_{\mathcal{U}_j} p(u_j|\hat h(u_j)) du_j	
\end{equation}
defined in terms of the marginal density $p(u_j|h)$ and a function $\hat h(u_j)$ that selects the maximal member in $\mathcal{M}_{RBM,j} = \{p(u_j|h) \,|\, h \in \mathcal{H}\}$ for every $u_j \in \mathcal{U}_j$. In general, $\hat h(u_j)$ will be different from $\hat h(u_j, \cdot)$ obtained for the joint distribution. 

Note that the estimation capacity of a model class $\mathcal{M}$ is computed without seeing any data. Figure 2 illustrates a distribution of Gaussians obtained in the decoder for different values of $h \in \mathcal{H}$. We show the marginals associated with one of the input dimensions. $C_{RBM,j}$ in equation(\ref{eq:cTj}) corresponds to the area below the ridgeline (in blue). Since all Gaussians have the same width $\sigma$ the area is maximized if the centers are distributed uniformly over an interval of interest. This allows us to define a simple regularization term.

In the following, we assume that the data has been normalized (e.g., transformed into $z$-scores) such that both variables have the same variance. Let $L_j = [k_j, l_j]$ be the range of values assumed by the $j$th component of the data, $j=1,2$. Further, let $\mathcal{U}_j^* = \{u_j^*(h)|h \in \mathcal{H}\}$ be the center set of the marginals over $u_j$. Let $\{\tau\}$ be an index set that sorts $\mathcal{U}_j^*$ in ascending order $u_j^*(\tau-1) \leq u_j^*(\tau)$, $\tau=2,\dots, 2^m$.

\textbf{Definition 3:} The function $d:\mathcal{U}^*_j \rightarrow \mathbb{R}_0^+$ is a measure of non-uniformity of the AE in the $j$th variable defined as
\begin{equation}
\label{eq:di}
d(\mathcal{U}_j^*) = \sum_{\tau=2}^{2^m} \left( u_j^*(\tau) - u_j^*(\tau-1) - \delta_j\right)^2  
\end{equation} 
for $\delta_j = (l_j-k_j)/2^m >0$ independent of $\tau$ and $u_{j}^*(1) \geq k_j$ and $u_{j}^*(2^m) \leq l_j$.

The higher $d(\mathcal{U}_j^*)$ is the lower the estimation capacity of the RBM with respect to the $j$th coordinate. Let $\mathcal{X^*}=\mathcal{U}_1^*$ be the location of the $x$-coordinates and $\mathcal{Y^*}=\mathcal{U}_2^*$ the location of the $y$--coordinates of the Gaussian centers. The term 
\begin{equation}
\label{eq:r2}
R = d(\mathcal{X}^*)+d(\mathcal{Y}^*)
\end{equation} 
attempts equal mode placement in both coordinates. At the same time, a small reconstruction error can only be achieved if the modes of the Gaussians lie on a manifold $\mathcal{D}$ 
\begin{equation}
	\label{eq:maniff}
	\mathcal{D} = \{(x^*,y^*) \in L \,| \,g(x^*,y^*) = 0\}
\end{equation} 
where $L = L_1 \times L_2$ and $g$ is a continuous function that defines the manifold. All observed data points $(x,y)$ are contained in a tubular neighborhood of $\mathcal{D}$ whose size scales with the reconstruction error. The AE needs to assign high data likelihood to points near $\mathcal{D}$ while at the same time attempting to place the modes uniformly along both coordinates. The following proposition states that this will almost surely fail for one of them.
\begin{figure*}[!htb]
\vskip 0.2in
\begin{center}
	\centerline{\includegraphics[width=1\columnwidth]{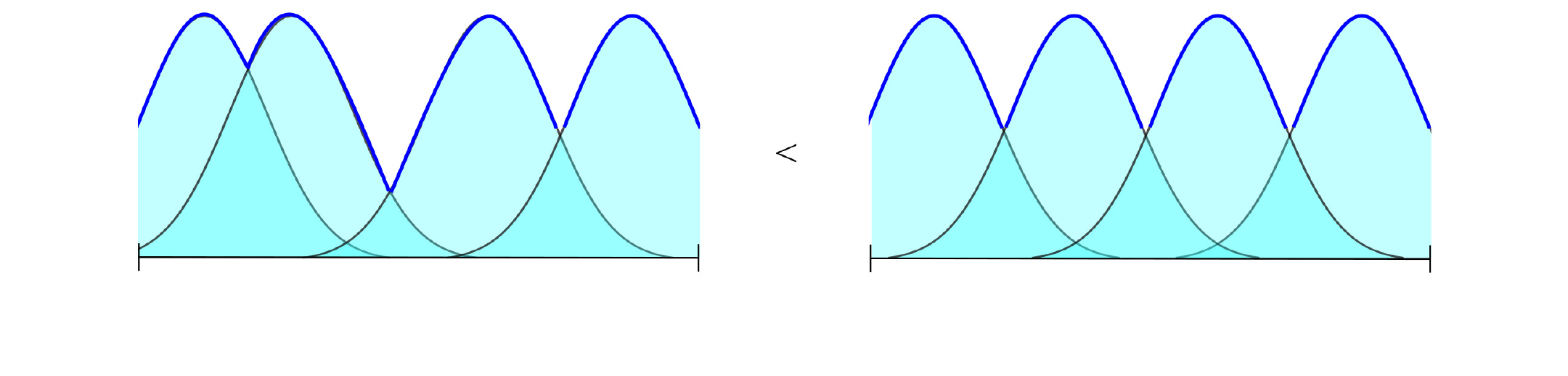}}
	\caption{Estimation capacity (area below blue ridge-line) as a function of mode placement of the Gaussians}
	\label{fig:roc}
\end{center}
\vskip -0.2in
\end{figure*}

\textbf{Proposition 1 (Structure identifiability):} Let $P_{X,Y}$ admit a uniform AE in $X$ with reconstruction error $\varepsilon_X$. Then, almost surely, it does not admit a uniform AE in $Y$ with reconstruction error $\varepsilon_Y \leq \varepsilon_X$.

\textbf{Proof:}
Being uniform in $Y$ without increasing the reconstruction error means that $\mathcal{D}$ supports uniformity in both $X$ and $Y$. This requires a particular form of $g(x^*,y^*)$ in equation (\ref{eq:maniff}): 
\begin{equation}
	\label{eq:smm}
	\begin{array}{rlll}
		g(x^*,y^*) &=& k_2 + a (x^*-k_1) -y^* = 0 \  \mbox{or}\\
		g(x^*,y^*) &=& l_2 - a (x^*-k_1) -y^*= 0\\
	\end{array}
\end{equation}
where $a = (l_2-k_2)/(l_1-k_1)$. $\mathcal{D}$, in turn, is generated by nature who selects $g$ randomly from the space $G$ of continuous functions on $\mathbb{R}\times\mathbb{R}$. The choices in (\ref{eq:smm}) form a set of measure zero in $G$. \hfill $\Box$

The regularized RBM (referred to as \textit{cRBM}) enables us to formulate a new criterion for causal inference.

\textbf{Criterion}: Let
\begin{equation}
	\label{eq:gamm}
	\gamma := d(\mathcal{X}^*) -d(\mathcal{Y}^*)
\end{equation}
If $\gamma <0$ we conclude $X\rightarrow Y$ and $Y\rightarrow X$ otherwise. If $\gamma=0$, no statement can be made. 

\section{Relation to IGCI}

We have introduced a criterion for bivariate causal discovery based on the idea that regularization works better for the cause than the effect variable. This viewpoint has some similarity to information-geometric causal-inference (IGCI) which we now explore. 

IGCI builds on the idea that the mechanism of selecting a distribution of the cause and the one translating the cause into the effect are independent. The idea is formalized in information space (i.e. using information-theoretic expressions) as an orthogonality condition between a function describing the local properties of the conditional $P_{Y|X}$ at a point $X=x$ and $P_X$. In its most common form, it is assumed that $Y = f(X)$ where $f$ is a smooth invertible function defined on $[0,1]$ that is strictly monotonic and satisfies $f(0) = 0$ and $f(1) = 1$. It can be shown that $\mbox{cov}(\log f^{\prime}, p_X) = 0$ implies that $\mbox{cov}(\log f^{-1^{\prime}}, p_Y) \geq 0$ where all quantities are considered as random variables \citep{janzing2012information}. Put differently, both directions cannot simultaneously be independent of the causal mechanism.

Contrasting this with the cRBM criterion (\ref{eq:gamm}) we see that it is also built on mutual exclusivity: if the AE of the joint data is uniform in $X$ it cannot at the same time be uniform in $Y$. If the AE is forced to make a decision it will find that regularization of the cause is more compatible with the data. The fact that one of the marginal densities has less structural constraints than the other motivates a regularization term which incorporates this fact as prior knowledge. As in the IGCI, the resulting criterion is not a statistical independence test of some kind but it is based on the geometry of the reconstruction. Mode placement in the decoder is a new way of decomposing distributions and understanding their structure. Notice that all quantities connected to mode placement are defined in terms of the set of \textit{available} activations $\mathcal{H}$, not the set of \textit{actual} activations $\{h^t\}, t=1,\dots,T$ generated for the $T$ observations. Only the geometry of their locations, not their occurrence in the given data-set is relevant.  

In the IGCI the mechanism is assumed to be deterministic and invertible. The cRBM imposes much weaker assumptions on the nature of the relationship between the two variables as expressed in (\ref{eq:maniff}). The main assumption is that the AE of the given data-set generalizes well in the sense that the reconstruction error obtained for held-out data (in a cross-validation procedure) is as low as the error obtained on training data. It is interesting to see that this enables us to identify the causal relation between $X$ and $Y$ which describes the behavior of the system under interventions, not just for new data from the same distribution. The clue is that our characterization of independence is aligned with improving the out-of-sample behavior of the model since it also constitutes a piece of prior knowledge: the fact that one of the variables is the cause 
implies uniformity as discussed above which is a \textit{useful} constraint on the geometry of the AE.

\section{Experiments}

We evaluate our method on both simulated data (for which the true causal direction is known) as well as real-life data. For the simulated data we use the data-sets discussed in \citep{mooij2016distinguishing} and create an additional new data-set for the case of a linear relationship between cause and effect. The real-life data is taken from the Tuebingen database of cause-effect pairs presented in the same paper and also contains data from the UCI Machine Learning Repository \citep{Dua:2019}.

\begin{table}[t]
	\caption{Choices of parameters for the different data sets}
	\label{sample-table}
	\vskip 0.15in
	\begin{center}
		\begin{small}
			\begin{sc}
				\begin{tabular}{lccccr}
					\toprule
					Parameter & \textsc{cep} & \textsc{sim-c} & \textsc{sim-lin} & \textsc{sim} \\
					\midrule
					$m$       &5&5&5&5\\
					$\sigma$  &.5&.5&.5&.5\\
					$\lambda$ & 1& 1& 1&3 \\
					\midrule
					$\eta$    &.001&.001&.001&.001\\
					$q$      &.9&.9&.9&.9\\
					\bottomrule
				\end{tabular}
			\end{sc}
		\end{small}
	\end{center}
	\vskip -0.1in
\end{table}

The simulated data are generated from a structural causal model (SCM) of the form 
\begin{equation}
	\label{eq:scm}
	\begin{array}{rrl}
		X&:=& N_1\\
		Y&:=& f(X, N_2)\\
	\end{array}
\end{equation}
where $f$ is a random function sampled from a Gaussian process and $N_1$, $N_2$ have random distributions, see \citep{mooij2016distinguishing}, appendix C.2, for details. In this paper, we discuss the datasets \textsc{sim}, the default case, \textsc{sim-lin}, the newly generated data-set in which $f$ is a linear function, \textsc{sim-c} which includes a confounding variable, and \textsc{cep}, the real-world data.

\begin{figure}[ht]
	\begin{center}
		\centerline{\includegraphics[width=0.45\columnwidth]{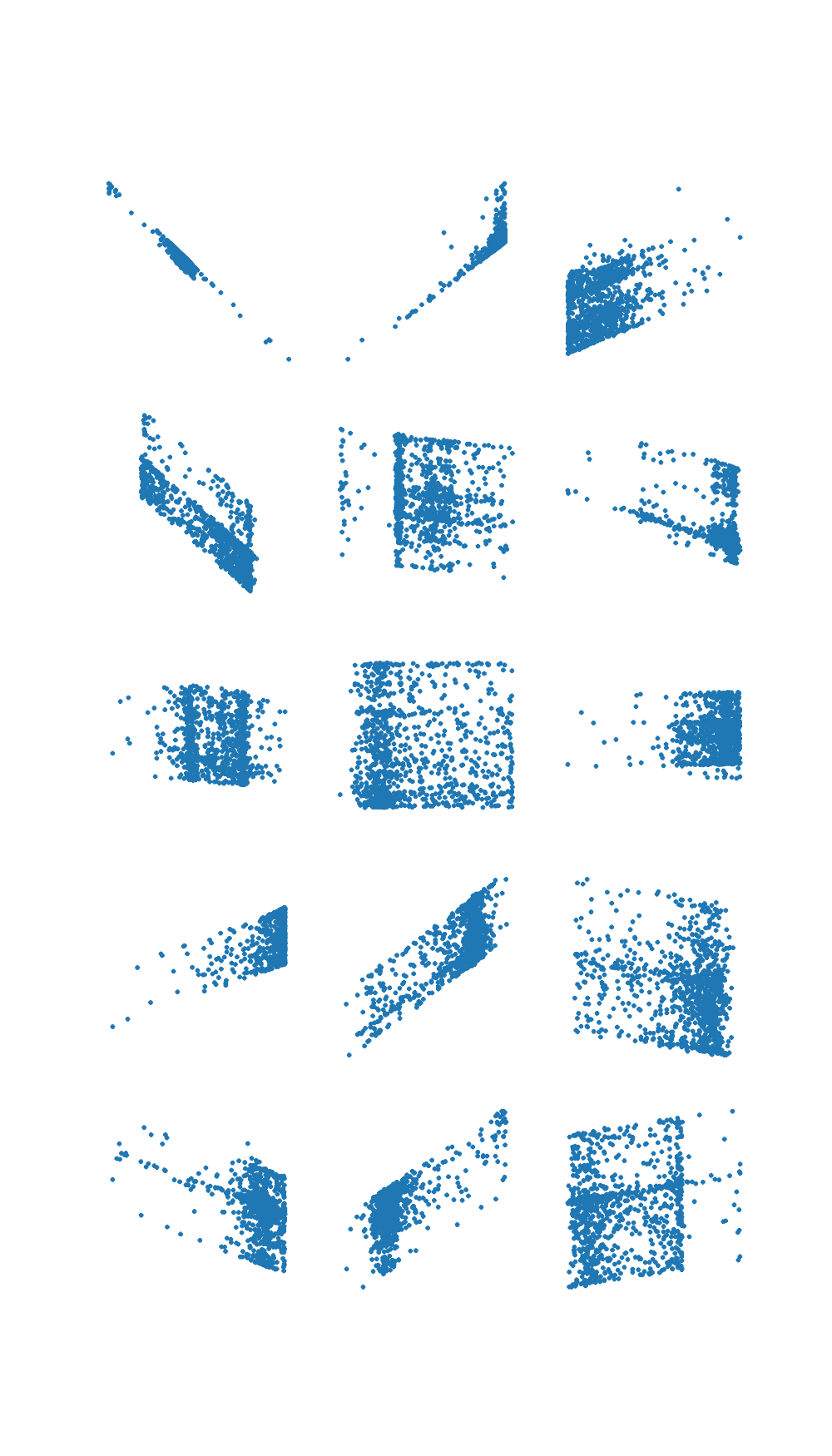}}
		\caption{Elements of the dataset \textsc{sim-lin}: scatter plots of simulated cause-effect pairs using a linear SCM (\ref{eq:sim-lin})}
		\label{fig:simlin}
	\end{center}
	\vskip -0.2in
\end{figure}

The new dataset \textsc{sim-lin} is generated using the following structural assignments
\begin{equation}
	\label{eq:sim-lin}
	\begin{array}{rll}
		X &:=& N_1\\
		Y &:=& bX + cN_2
	\end{array}
\end{equation}
where $N_1, N_2$ are sampled from random distributions, $b\sim U[-1,1]$ is the slope parameter sampled from a uniform distribution and $c$ is chosen in a way that ensures $\mbox{var}(Y) = \mbox{var}(X)$ such that the slope remains unchanged when normalizing the data. Every dataset consists of 100 cause--effect pairs and 1000 observations each. Figure (\ref{fig:simlin}) displays a few examples of pairs in \textsc{sim-lin}. 

The dataset \textsc{CEP} is a collection of 108 real-world variables for which the true causal direction is technically unknown but a fundamental analysis of the supposed causal mechanism provides strong arguments for the assigned true direction. These arguments are detailed in appendix D of \citep{mooij2016distinguishing}. Some of the ``pairs'' consist of multi-dimensional variables which we project on their 1st principal component before proceeding. Another pre-processing step is to normalize all data by computing the z-score on both  $x_1,\dots x_T$ and $y_1,\dots y_T$.

A Gaussian-Bernoulli RBM as introduced in section 2 is trained on all datasets. The critical parameter in the RBM is the number $m$ of hidden units which determines the number of available activations. It also defines the dimensions of the weight matrix and bias vectors and, hence, the number of tunable parameters. The shape parameter $\sigma$ in (\ref{eq:back}) is kept fixed. The RBM becomes a cRBM (RBM for causal discovery) by including the regularization term (\ref{eq:r2}) into the loss function. We do this using the scalar weight $\lambda>0$ and obtain 
\begin{equation}
	\label{eq:lf}
	L(\theta) = \mbox{CD}(\theta) + \lambda R(\theta) 
\end{equation}
where $CD$ refers to contrastive divergence discussed in section 2. $L(\theta)$ is minimized w.r.t. $\theta$ using gradient descent. We use stochastic gradient descent which has the desirable property that the parameters converge to flat regions of the loss function. This means that the numerical values of the optimal $\theta$ can be truncated after a few digits as a small perturbation of the weights (due to rounding) does not alter the RBM performance. For every pair, training is performed over 5000 epochs and stopped prematurely if the loss function stops decreasing. Training typically stops after 1000-2000 epochs in the experiments reported here. An important meta-parameter is the step-size $\eta$ of the gradient. A pre-factor $q<1$ imposes an exponential decay of the step-size over epochs to avoid oscillations near a (local) minimum in the loss function. The cRBM is implemented using Tensorflow \citep{tensorflow2015-whitepaper} and is available as supplementary material to this paper. The choices of the (meta-) parameters for the different experiments are reported in table 1.

\begin{figure*}[!htb]
	\vskip 0.2in
	\begin{center}
		\centerline{\includegraphics[width=1\columnwidth]{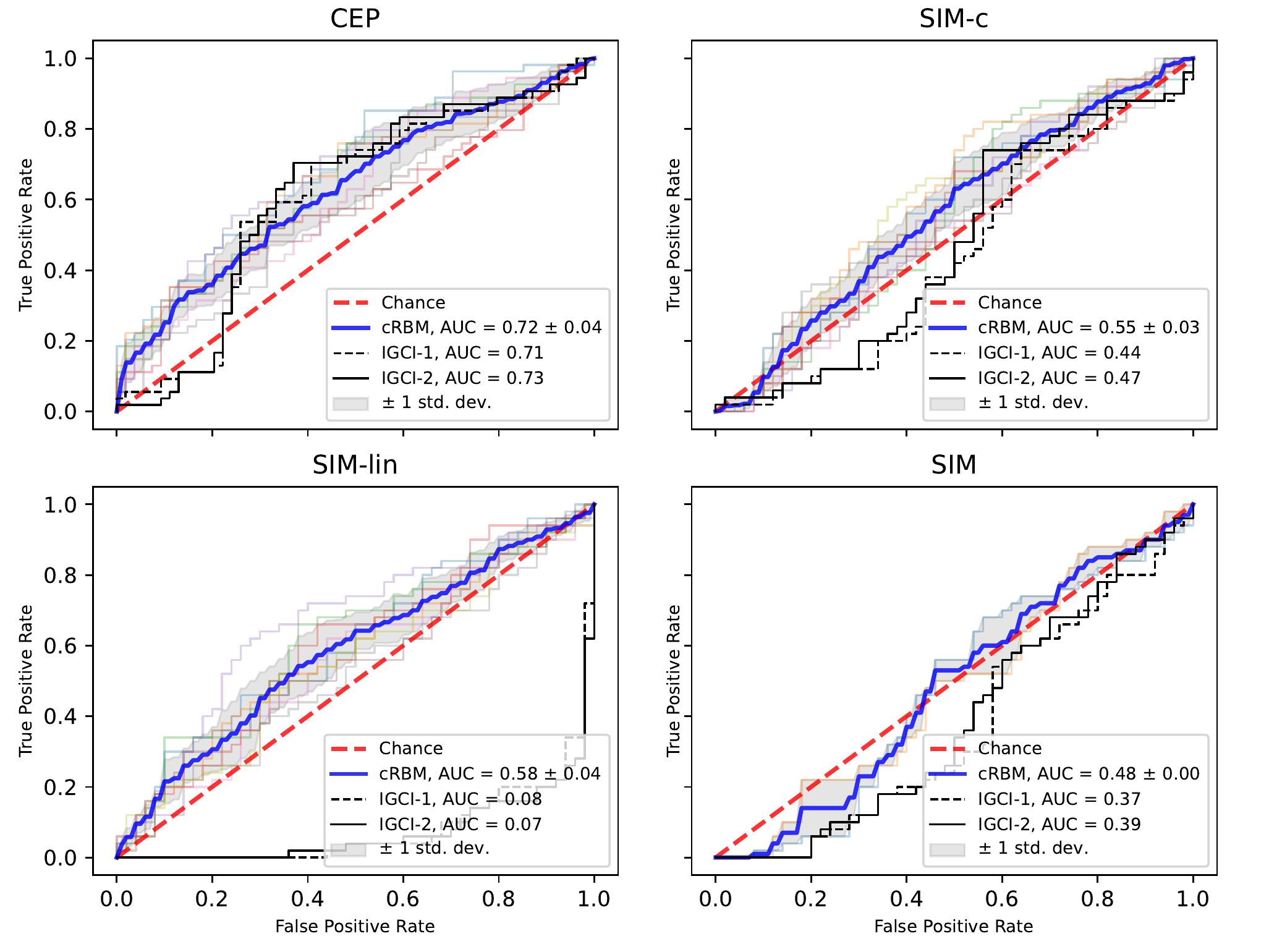}}
		\caption{ROC curves and AUCs obtained for the different datasets}
		\label{fig:roc}
	\end{center}
	\vskip -0.2in
\end{figure*}

Performance is evaluated using two metrics, the accuracy and the area under the ROC curve of the cRBM. To construct the ROC curve we use $\gamma$ as defined in equation (\ref{eq:gamm}) as a heuristic estimate of confidence. As benchmarks we use IGCI-1 and IGCI-2, the slope and entropy based estimators of (information-geometric) independence, both using the uniform distribution as their reference measure. To obtain the first metric, our criterion is applied as defined in (\ref{eq:gamm}) and we simply count the number of correct decisions. For the \textsc{cep} data, pairs belonging to the same real-world experiment are weighted such that the entire experiment only counts as one. This is in accordance to the procedure applied in \citep{mooij2016distinguishing}. As the training of the cRBM is itself a stochastic procedure (in particular due to the sampling of activations $h$ as part of reproducing the input distribution), we report the mean and standard deviation obtained during multiple training rounds. The results are detailed in table 2. 

\begin{table}[t]
	\caption{Accuracies obtained for the different datasets}
	\label{sample-table}
	\vskip 0.15in
	\begin{center}
		\begin{small}
			\begin{sc}
				\begin{tabular}{lccc}
					\toprule
					& \textsc{crbm} & \textsc{igci-1} & \textsc{igci-2}\\ 
					\midrule
					\textsc{cep} & 0.64 $\pm$ 0.04&0.64&0.69\\
					\textsc{sim-c} & 0.54 $\pm$ 0.03&0.45&0.49\\
					\textsc{sim-lin}& 0.57 $\pm$ 0.04&0.18&0.15\\
					\textsc{sim} &0.51 $\pm$ 0.005&0.37&0.42\\
					\bottomrule
				\end{tabular}
			\end{sc}
		\end{small}
	\end{center}
	\vskip -0.1in
\end{table}

The ROC-curves are displayed in figure (\ref{fig:roc}) which also contains the AUCs obtained for the different datasets. Interestingly, the performance of cRBM on real data \textsc{cep} is comparable to the IGCI which may be testament to the fact that the two methods ultimately seek to exploit the same property, namely that the distribution of the cause can be chosen independently of the mechanism responsible for the effect. However, the techniques for extracting this property from data are quite different: the cRBM tries to solve a data reproduction problem through ``mode placement'' in the decoding function, while the IGCI tries to establish orthogonality between information theoretic quantities estimated from the data. Mode placement seems to be more universal as it almost always fails for one of the candidate variables, as outlined in proposition 2. It is clearly superior for the simulated data with confounder \textsc{sim-c} but stays close to pure chance level for the basic \textsc{sim} data which is somewhat surprising but not different from IGCI. A potential explanation is that despite being generated from an SCM with random functions, the \textsc{sim} data is still too ``obvious''. It is well known that additional sources of noise can help finding global minima in rugged loss landscapes but a first attempt to improve performance by adding Gaussian observation noise failed to improve performance. The latter is quite satisfactory for the newly generated data \textsc{sim-lin} where the mechanism is linear. As expected, the IGCI benchmark fails in this case as the (linear) mechanism does not create any footprint in the distribution of the effect.

\section{Conclusion}
The paper introduces a new way of identifying the causal direction among two associated variables based on the postulate of independent mechanisms. The idea is to find a representation of their joint distribution by means of an auto-encoder and to realize that its performance can be improved by incorporating the condition that one of the marginal densities is free of structural constraints. This is achieved by adding a regularization term to the loss function of the auto-encoder. The regularization targets the location of the modes of the decoder density functions and seeks to distribute them uniformly within a region of interest which represents an intuitive, geometric criterion. Given the importance of autoencoders in unsupervised machine learning, the paper also contributes to the search for regularization techniques that unveil structural (in our case causal) relations among the data.

\vskip 0.2in
\bibliography{mybib}

\end{document}